\documentclass{article} 
\usepackage[preprint]{colm2025_conference}

\usepackage{microtype}
\usepackage{hyperref}
\usepackage{url}
\usepackage{booktabs}
\usepackage{pifont}
\usepackage{bm}
\usepackage{bbm}
\usepackage{amsfonts}       
\usepackage{nicefrac}       
\usepackage{xcolor}         
\usepackage{wrapfig}
\usepackage{multirow,makecell}
\usepackage{tabularx}
\usepackage{subfigure}
\usepackage{bbding}
\usepackage{threeparttable}
\usepackage{wasysym}
\usepackage{todonotes}
\usepackage{enumitem}
\usepackage{makecell}
\usepackage{xspace}
\usepackage{caption}
\usepackage{subcaption}

\usepackage{lineno}

\definecolor{darkblue}{rgb}{0, 0, 0.5}
\hypersetup{colorlinks=true, citecolor=darkblue, linkcolor=darkblue, urlcolor=darkblue}

\newcommand{\alg}{\textsc{ThinkPrune}\xspace}

\title{{\alg}: Pruning Long Chain-of-Thought of LLMs via Reinforcement Learning}

\author{Bairu Hou$^{1}$\quad \quad \quad Yang Zhang$^2$\quad \quad \quad Jiabao Ji$^1$\quad \quad \quad Yujian Liu$^1$\quad \quad \quad Kaizhi Qian$^2$ \\ \textbf{Jacob Andreas$^3$\quad \, Shiyu Chang$^1$} \\
\\
  $^1$UC Santa Barbara \quad \quad \quad \quad
  $^2$MIT-IBM Watson AI Lab \quad \quad \quad \quad
  $^3$MIT CSAIL \\
}
\begin{document}

\ifcolmsubmission
\linenumbers
\fi

\maketitle

\begin{abstract}
We present {\alg}, a simple yet effective method for pruning the thinking length for long-thinking LLMs, which have been found to often produce inefficient and redundant thinking processes. Existing preliminary explorations of reducing thinking length primarily focus on forcing the thinking process to early exit, rather than adapting the LLM to optimize and consolidate the thinking process, and therefore the length-performance tradeoff observed so far is sub-optimal. To fill this gap, \alg offers a simple solution that continuously trains the long-thinking LLMs via reinforcement learning (RL) with an added token limit, beyond which any unfinished thoughts and answers will be discarded, resulting in a zero reward. To further preserve model performance, we introduce an iterative length pruning approach, where multiple rounds of RL are conducted, each with an increasingly more stringent token limit. We observed that \alg results in a remarkable performance-length tradeoff --- on the AIME24 dataset, the reasoning length of \texttt{DeepSeek-R1-Distill-Qwen-1.5B} can be reduced by half with only 2\% drop in performance. We also observed that after pruning, the LLMs can bypass unnecessary steps while keeping the core reasoning process complete. Code is available at \url{https://github.com/UCSB-NLP-Chang/ThinkPrune}.
\end{abstract}

\section{Introduction}
Recent advances in large language models (LLMs) have demonstrated the effectiveness of inference-time scaling through reinforcement learning~\citep{deepseek-ai2025deepseekr1, openai2024openai, code-r1, zeng2025simplerl0zoo0}, where LLMs learn to produce long and sophisticated reasoning behaviors such as self-refection and verification, significantly increasing their performance on a wide range of benchmarks. However, one key challenge of inference-time scaling is the significant number of tokens produced at inference time, leading to high computational and memory overhead. For example, on the MATH500~\citep{lightman2023math500} benchmark, the \texttt{DeepSeek-R1-Distill-Qwen-1.5B} model generates solutions with more than 15,000 tokens on average, while many of the questions could have been solved with fewer than 1,000 tokens by regular LLMs. This highlights the issue of \emph{over-thinking}, where many reasoning steps might be redundant or inefficient~\citep{kumar2025overthink, chen2024not, sui2025stop}.

\begin{figure}[t]
    \centering
    \hspace{-2mm}
    \includegraphics[width=\linewidth]{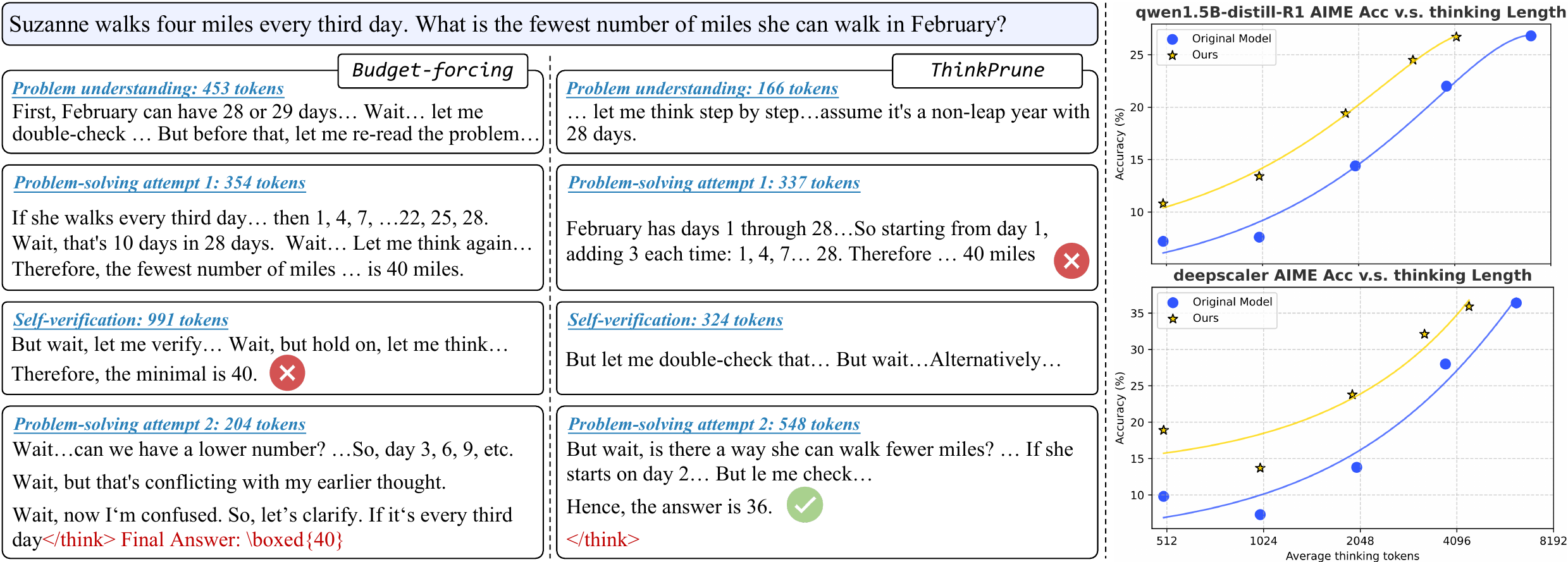}\\
    \quad \quad \quad \quad \quad\quad \quad {\small (a) Example Reasoning Chains.} \quad  \quad \quad \quad \quad \quad \quad \quad{\small (b) Accuracy vs. Think Length.}
    \caption{Comparison between budget forcing and \alg. (a) Example Reasoning Chains: Under a 2000-token thinking budget, applying budget-forcing on the original model uses up all token budgets before identifying the mistake, leading to a wrong answer. In contrast, the model trained with {\alg} solves the problem more efficiently, using fewer tokens and giving the correct answer. (b) Performance of the original LLM and the model after {\alg} training under different thinking token budgets.}
    \label{fig: intro}
\end{figure}

There are now some preliminary explorations of limiting the generation length via \textit{budget-forcing}~\citep{fu2024efficiently, muennighoff2025s1}, where, when reaching a given token limit, the thinking process is forced to early exit, \emph{e.g.}, by appending an end-of-thinking token and producing an answer right away. Figure~\ref{fig: intro}(a) (left) shows an example output of S1 \citep{muennighoff2025s1}, a budget-forcing method, under a low budget (2000 thinking tokens), where the thinking process spent 453 tokens just to understand the problem, and therefore could not complete its second round of thought within the budget. Accordingly, a non-trivial performance drop would occur as the budget gets tighter, as shown in Figure~\ref{fig: intro}(b) (blue lines).

However, this is apparently sub-optimal. When the budget is low, a much more sensible solution is to remove the redundant thinking or unimportant steps (such as the problem paraphrasing), rather than maintaining the inefficient thinking and getting it killed. As a result, the performance drop as thinking length shrinks may have been significantly overestimated. So far, there have not been explorations that seek to fine-tune the model to adapt to smaller thinking budgets, and thus, we still do not have a good estimate of how much redundancy there is in the long thinking process that can be removed. Specifically, the following research questions remain unanswered:
\setlength{\leftmargini}{10pt}
\begin{itemize}
    \item Can we fine-tune an LLM with long CoT to prune its thinking length, while minimizing the performance drop?
    \item What would be the length-performance tradeoff when the long CoT is pruned?
    \item What happens to the reasoning chain when it gets pruned? What steps or words are most likely to be pruned?
\end{itemize}

In this paper, we seek to fill this gap. We propose \alg, a simple yet effective length-pruning strategy for LLMs with long CoT, which enforces a maximum generation length during RL training. Specifically, given an LLM with long CoT, we perform continuous RL following the same scheme of DeepSeek-R1, except that we impose a strict token limit during training (\emph{e.g.}, 4,000 tokens for both reasoning and answer tokens). Any tokens beyond this limit are discarded before reward computation. This means that even if the model generates a correct answer beyond the allowed length, it still receives a reward of 0 because the answer is clipped and cannot be extracted. The task performance can be further preserved with an iterative pruning strategy, where multiple rounds of RL are conducted with increasingly more stringent length limits.

Extensive evaluation demonstrates a strong trade-off between generation length and performance for \alg. For example, {\alg} reduces the average generation length from 10,355 to 3,574 tokens for the \texttt{DeepSeek-R1-Distill-Qwen-1.5B} model, while also improving average accuracy across four math benchmarks. Although there is a slight performance drop of 2\% for the \texttt{DeepScaleR-1.5B-Preview} and \texttt{QwQ-32B} models, our method achieves a comparable reduction in generation length --- from 5,914 to 3,370 tokens, and from 8,763 to 4494, respectively. 
Figure~\ref{fig: intro}(b) (yellow lines) also shows the improved accuracy-thinking length tradeoff.
Further analysis shows that \alg helps LLMs avoid unnecessary reasoning steps while maintaining focus on solving the question, as shown in Figure~\ref{fig: intro}(a) (right), where problem understanding drops to 166 tokens.
These findings provide valuable insights into improving the inference-time reasoning efficiency for long-COT LLMs.

\section{Related Work}

\subsection{Reinforcement Learning for LLM Reasoning}
Reinforcement learning (RL) has shown strong potential in improving the reasoning abilities of LLMs across various domains, such as math~\citep{deepseek-ai2025deepseekr1} and coding~\citep{openai2025competitive, code-r1}, and complicated browser surfing~\cite{openai_operator_2024}. The resulting long-COT LLMs such as OpenAI-o3~\citep{openai2024openai} and DeepSeek-R1~\citep{deepseek-ai2025deepseekr1} significantly outperform short-COT LLMS and demonstrate that reinforcement learning with verifiable reward (RLVR) can encourage LLMs to develop deep thinking behaviors, such as broad exploration and feasibility checks~\citep{gandhi2025cognitive}, without relying on complex reasoning data generation methods like Monte-Carlo Tree Search~\citep{zelikman2024quiet, hosseini2024vstar}. 
However, these behaviors often lead to much longer reasoning traces, sometimes several times longer than those produced by short COT LLMs~\citep{sui2025stop, chen2025towards}, creating an ``overthinking'' issue that largely increases inference costs~\citep{kumar2025overthink}. 
Several recent works have shown that this extended reasoning often includes redundant or unnecessary verification and reflection, even on simple problems~\citep{wang2025thoughts, ji2025first}. 
Our work follows the standard RL training pipeline without changing the reward function and shows that it is possible to retain strong reasoning performance while significantly reducing overthinking.

\subsection{Efficient Long Chain-of-Thought LLM}
Several works have aimed to improve token efficiency in long chain-of-thought (CoT) LLMs. For example, \citet{team2025kimi}, \citet{chen2024not}, and \citet{shen2025codi0} reduce the length of reasoning by adding a length penalty during RL training. Other works, such as \citet{hao2024training} and \citet{geiping2025scaling}, represent reasoning as an optimization over latent vectors instead of text tokens, which helps shorten the reasoning process.
While these methods are effective when training long CoT LLMs from short COT ones, there have been few works on reducing the reasoning length for trained LLMs. The most relevant works to ours are test-time methods that shorten reasoning with early exit strategies~\citep{muennighoff2025s1, fu2024efficiently, zeng2025revisiting}. These methods add stopping tokens or limit the maximum number of generated tokens during testing. However, they often lead to significant performance drops, especially when early stopping occurs too soon in the reasoning process.
In contrast, our work presents a simple and effective method with further RL training to reduce reasoning length without sacrificing performance.

\section{Method}

\subsection{Overview}
Denote $\mathcal{M}_{\bm \theta_0}(\cdot)$ as an LLM capable of performing long CoT. Given a query $\bm q$, a sample answer, along with a long reasoning chain, is sampled from the LLM's output distribution, \emph{i.e.}, $\bm Y \sim \mathcal{M}_{\bm \theta_0}(\bm q)$. Our goal is to fine-tune the LLM, such that its output length is reduced while the overall performance is desirably maintained.

{\alg} is a simple yet effective RL strategy to achieve the length reduction goal. In what follows, we will first introduce the RL framework of {\alg}, and then introduce an iterative length reduction strategy to better maintain the performance.

\subsection{Reinforcement Learning with Length Clipping}
\label{subsec:rl}
{\alg} adopts a similar RL scheme to the DeepSeek-R1 model~\citep{deepseek-ai2025deepseekr1} while reducing the generation length. Specifically, we adopt the group relative policy optimization (GRPO) algorithm \cite{shao2024deepseekmath}. The reward function is almost the same as the DeepSeek-R1 framework, \emph{except} that a \emph{length clipping} is added. Formally, denote $L$ as the length limit. The reward function is defined as follows:
\begin{equation}
    R(\bm Y, \bm q; L) = \left\{
    \begin{array}{cl}
        1 &  \mbox{ if an answer can be extracted from clip}(\bm Y, L) \mbox{ and is correct,}  \\
        0 & \mbox{ otherwise;} 
    \end{array}
    \right.
\end{equation}
where clip$(\bm Y; L)$ represents clipping the output $\bm Y$ to length $L$. In other words, the only difference from the DeepSeek-R1 reward is that the model-generated output is clipped to $L$ before the reward is evaluated. In this way, output with length above $L$ would not be able to produce a valid answer since it is cut off, thus receiving a zero reward. This clipping operation effectively encourages the model to produce answers below the length limit. 
Such reward design is very simple, so it does not involve any hyperparameter tuning or reward engineering, and can ensure training stability. Also, since it involves minimal tweaks to DeepSeek-R1's training strategy, the proposed solution helps to maintain the long reasoning capabilities inherent in the base model.

During training, we also append a simple system prompt into each training example to explicitly tell the model the length limit, such as \textit{The output of the assistant should be within L tokens}, to explicitly tell the model the length limit. The full system prompt for each model is shown in Appendix~\ref{app: implementation}.

\subsection{Iterative Length Pruning Strategy}\label{subsec: pruning-strategy}

The success of the proposed algorithm relies on the choice of the length limit, $L$ --- if $L$ is set too stringently compared to the original output length of the base model, then the task performance can be seriously compromised.

Drawing inspiration from the iterative pruning strategy for reducing model parameters \citep{han2015learningweightsconnectionsefficient}, we propose an iterative length pruning scheme. Denote $L^*$ as the target length constraint, we introduce a length schedule, $L_1 > L_2 > \cdots > L^*$. At each iteration $t$, we reduce the length constraint to $L_t$, and further fine-tune the model $\bm \theta_{t-1}$ from last iteration to $\bm \theta_t$, using the RL procedure described in Section~\ref{subsec:rl}. Such an iterative pruning strategy ensures that the LLM learns to recover the performance by gradually compacting its output reasoning chain.

A critical design choice of the iterative length pruning is the stopping criterion for each iteration of the RL training. In our implementation, we utilize AIME22 and AIME23 as the dataset as a validation set to choose the best checkpoint for the next RL iteration. To better balance model performance and generation length reduction, we allow up to a relative 10\% drop in pass@1 accuracy on AIME-22 and AIME-23 compared to the original model. Among the checkpoints that meet this criterion, we select the one with the shortest average output length as the starting checkpoint.

\section{Experiment}
In this section, we conduct empirical evaluations to assess the effectiveness of our proposed method. We first present the experiment setup in Section~\ref{sec:experiment-setup}. Then, we present the experiment results in Section~\ref{sec:main-experiment}.

\subsection{Experiment Setup}\label{sec:experiment-setup}
\paragraph{Backbone models.}
Representative open-sourced long reasoning LLMs include DeepSeek-R1~\citep{deepseek-ai2025deepseekr1} and QwQ~\citep{qwq32b}, along with their distilled variants. In our experiments, we use three models from these families: Distill-Qwen-1.5B, DeepScaleR-1.5B-Preview, and QwQ-32B.

We group these models into two categories based on the extent of their training: \textit{unsaturated} and \textit{saturated} models.
Specifically, while Deepseek-R1-Distill models have been widely used, these models are directly fine-tuned on the output of DeepSeek-R1 in a supervised fine-tuning manner instead of RL, which could limits their full potential. Previous work, such as \citet{deepscaler2025}, has shown that further training these models with RL can improve their performance, especially on math benchmarks.
Based on this, we treat Distill-Qwen-1.5B as an unsaturated model and DeepScaleR-1.5B-Preview as its saturated version, since it is further trained with RL. Similarly, QwQ-32B is also trained with RL and is considered a saturated model.
This selection of models, covering both unsaturated and saturated types, allows us to more thoroughly evaluate the effects of thinking length pruning.

\paragraph{Training datasets.}
Previous work~\citep{ye2025limo} has shown that even a small but high-quality training dataset can improve the LLM performance via RL. Therefore, we utilize a small number of data for model training by only using the historical AIME and AMC math questions~\citep{AoPS_AIME}. We use the preprocessed data from Prime~\citep{cui2025prime} and take the AIME-AMC subset for training. In total, the training dataset consists of 2470 distinct training examples.

\paragraph{Comparisons.} We compare with the original backbone models without pruning. Additionally, we include the budget-forcing method~\citep{muennighoff2025s1} for length reduction, which enforces a maximum number of thinking tokens by appending the end-of-thinking token delimiter (detailed implementation is described in Appendix \ref{app: implementation}).
For our method, we report the following two variants:
\begin{itemize}[leftmargin=10pt,itemsep=2pt, topsep=2pt]
\item One-shot length pruning: We set the maximum length to 4,000/3,000/2,000 tokens respectively and directly perform RL to reduce the generation length of the LLM.
\item Iterative length pruning: Starting from a higher generation length budget, we perform multi-round RL training and gradually decrease the maximum length after each iteration.
\end{itemize}

\paragraph{Implementation details.} We use the Verl~\citep{sheng2024verl} RL framework for high-performance training. All models are trained with a batch size of 128. The number of rollouts for each question is set to 16 following prior works~\citep{zeng2025simplerl0zoo0, code-r1}. We use the GROP algorithm \citep{shao2024deepseekmath}. For both one-shot and iterative pruning, we employ the same checkpoint selection strategy as mentioned in Section~\ref{subsec: pruning-strategy}, where the checkpoint with the shortest average output length while maintaining a relative 10\% accuracy drop is selected.

\paragraph{Evaluation configurations.} 
We follow prior works to include the following evaluation datasets: MATH-500~\citep{lightman2023math500}, AIME24~\citep{AoPS_AIME}, AMC23~\citep{maa_amc}, and OlympiadBench~\citep{he2024olympiadbench0}. We use the versions of these dataset hosted in the Qwen2.5-Math GitHub repository for ease of reference.
Following DeepSeek R1~\citep{deepseek-ai2025deepseekr1}, we set the maximum generation length (including both the thinking tokens and answer tokens) to 32,768 tokens for all the models, far above the maximum token length during our training. For each testing question, we sample $N$ outputs with a temperature of 0.6 and a top-p value of 0.95, and we report the average accuracy of these $N$ outputs. The number of samplings varies depending on the model size and dataset size. Specifically, for the two 1.5B models, we use $N=64$ for AIME24 and AMC23 and $N=16$ for MATH500 and OlympiadBench. This also refers to the evaluation hyper-parameters of DeepSeek R1, where the number of sampled responses are adjusted between 4 and 64 depending on the test set size to balance the variance and evaluation cost. For \texttt{QwQ-32B}, we sample $N=16$ responses for each question, given its large size and high computational cost.

Since the accuracy evaluation requires complex evaluation of mathematical expressions, we adopt the math evaluator from Qwen-2.5-math~\citep{deepseek-ai2025deepseekr1}, which provides robust answer extraction and advanced expression comparison.

\begin{table*}[t]
\centering
\resizebox{\textwidth}{!}{%
\begin{tabular}{l|ccccc|ccccc}
\toprule \midrule
& \multicolumn{5}{c|}{\textbf{Accuracy}} & \multicolumn{5}{c}{\textbf{Generation Length}} \\
& \makecell{MATH\\500} & \makecell{AIME} & \makecell{AMC} & \makecell{Olympiad\\Bench} & Avg. & \makecell{MATH\\500}  & \makecell{AIME} & \makecell{AMC} & \makecell{Olympiad\\Bench} & Avg. \\
\midrule
\multicolumn{11}{c}{\textbf{\texttt{DeepSeek-R1-Distill-Qwen-1.5B}}} \\
\midrule
Original Model & 82.9 & 29.4 & 70.3 & 44.7 & 56.8 & 5560 & 15484 & 10030 & 11526 & 10355 \\
{\alg}-4k & 83.8 & 29.0 & 73.6 & 46.5 & 58.2 & 2709 & 8301 & 4388 & 5529 & 5232 \\
{\alg}-3k & 83.7 & 27.8 & 71.8 & 44.9 & 57.1 & 2557 & 7968 & 4096 & 5140 & 4940\\
{\alg}-2k & 82.9 & 27.0 & 72.2 & 45.6 & 56.9 & 2356 & 7574 & 3755 & 4913 & 4650 \\
{\alg}-4k$\rightarrow$ 3k &  83.9 & 26.9 & 71.4 & 46.0 & 57.1 & 2209 & 6389 & 3422 & 4229 & 4062 \\
{\alg}-4k$\rightarrow$ 3k$\rightarrow$ 2k & 83.2 & 27.1 & 73.2 & 46.2 & 57.4 & 1938 & 5631 & 3039 & 3687 & 3574\\

\midrule
\multicolumn{11}{c}{\textbf{\texttt{DeepScaleR-1.5B-Preview}}} \\
\midrule
Original Model & 88.5 & 40.3 & 81.2 & 52.7 & 65.7 & 3084 & 9463 & 5202 & 5907 & 5914 \\
{\alg}-4k & 87.1 & 37.2 & 80.2 & 51.4 & 64.0 & 2212 & 6366 & 3516 & 4055 & 4037 \\
{\alg}-3k & 86.5 & 34.3 & 78.8 & 50.6 & 62.6 & 1991 & 5809 & 3122 & 3583 & 3626 \\
{\alg}-2k & 86.1 & 33.3 & 77.7 & 49.7 & 61.7 & 1880 & 5528 & 2961 & 3348 & 3429 \\
{\alg}-4k$\rightarrow$ 3k & 87.1 & 38.4 & 79.9 & 51.6 & 64.2 & 2086 & 5869 & 3278 & 3731 & 3741 \\
{\alg}-4k$\rightarrow$ 3k$\rightarrow$ 2k & 86.9 & 36.5 & 79.4 & 50.1 & 63.2 & 1881 & 5301 & 2963 & 3334 & 3370\\

\midrule
\multicolumn{11}{c}{\textbf{\texttt{QwQ-32B}}} \\
\midrule
Original Model & 95.1 & 78.8 & 97.5 & 71.1 & 85.6 & 4289 & 13822 & 7442 & 9497 & 8763 \\
{\alg}-4k & 94.0 & 76.3 & 95.8 & 68.2 & 83.5 & 2552 & 8787 & 4173 & 5687 & 5300 \\
{\alg}-3k & 94.0 & 75.0 & 95.8 & 68.6 & 83.3 & 2341 & 8308 & 3943 & 5413 & 5001 \\
{\alg}-2k & 93.5 & 73.3 & 95.5 & 68.6 & 82.7 & 2133 & 8232 & 3770 & 5160 & 4824 \\
{\alg}-4k$\rightarrow$ 3k & 94.0 & 73.8 & 96.1 & 67.6 & 82.9 & 2308 & 8176 & 3959 & 5301 & 4936 \\
{\alg}-4k$\rightarrow$ 3k$\rightarrow$ 2k & 93.8 & 72.5 & 95.9 & 67.3 & 82.4 & 2162 & 7631 & 3441 & 4742 & 4494\\
\midrule
\bottomrule
\end{tabular}%
}
\caption{Performance visualization of {\alg}. The accuracy is measured by sampling multiple responses from the LLMs and taking the average to reduce variance.}
\label{tab: main_exp}
\vspace*{-0.2in}
\end{table*}

\subsection{Main Experiment: Length Pruning}\label{sec:main-experiment}

\paragraph{One-shot length pruning enables significant thinking length reduction.} We first evaluate the effectiveness of one-shot pruning in Table~\ref{tab: main_exp} (top 3 rows in each section), where the LLMs are directly trained with a length limit 2k/3k/4k. We highlight the following observations.
First, the simple strategy can effectively reduce the generation length with moderate negative effect on the performance. For the unsaturated model, \texttt{DeepSeek-R1-Distill-Qwen-1.5B}, the length reduction rate can be up to 50\% with one-shot length pruning. At the meantime, the average performance is well-maintained and even slightly improved. For saturated models like \texttt{DeepScaleR-1.5B-Preview} and \texttt{QwQ-32B}, we observe a 40–50\% reduction in token usage, with moderate performance degradation. This highlights the promising efficiency gains through length pruning, especially when models are initially over-generating.
Second, we observe a consistent tradeoff between length and performance for all models under one-shot pruning. As we lower the token limit from 4k to 2k, the number of tokens goes down and the accuracy drops slightly. This suggests that cutting more reasoning tokens aggressively may also limit the reasoning capabilities of the LLMs.

Another interesting phenomenon is that even though the model is trained with an explicit length limit, it often goes beyond that limit at test time. Particularly, we observe that although trained with an explicit length limit, the model can still generate long responses when the problem becomes more difficult such as on the AIME dataset. This shows that our length pruning does not hurt the model's deep thinking behavior, and it maintains the ability to perform complex reasoning for difficult questions.

\paragraph{Iterative length pruning benefits performance preservation and length reduction.} 
The second question we aim to explore is whether iterative length pruning can better main the original LLM performance with a decent thinking length pruning compared to one-shot pruning. As shown in Table~\ref{tab: main_exp} (last two rows in each section), we start with LLMs trained using a 4k length limit and then iteratively apply RL with reduced length limits from 3k to 2k.
Our findings are as follows. 
\textbf{First}, for the two 1.5B models, iterative pruning leads to better performance with either shorter or similar response lengths compared to one-shot pruning under the same final length limit. For example, on \texttt{DeepScaleR}, {\alg}-4k$\rightarrow$3k$\rightarrow$2k outperforms the one-shot 2k model by 1.5\% in accuracy while using 59 fewer tokens on average. Similarly, pruning from 4k to 3k leads to better results than directly pruning to 3k.
\textbf{Second}, for the \texttt{QwQ-32B} model, performance drops slightly after length pruning, with an average decrease of 2.7\%. Unlike the 1.5B models, iterative pruning does not help recover the lost performance. We believe this is because it struggles to maintain performance when the generation token budget is too tight. We will discuss this in more detail in the next paragraph.

\begin{figure}[t]
\centering
\includegraphics[width=0.9\textwidth]{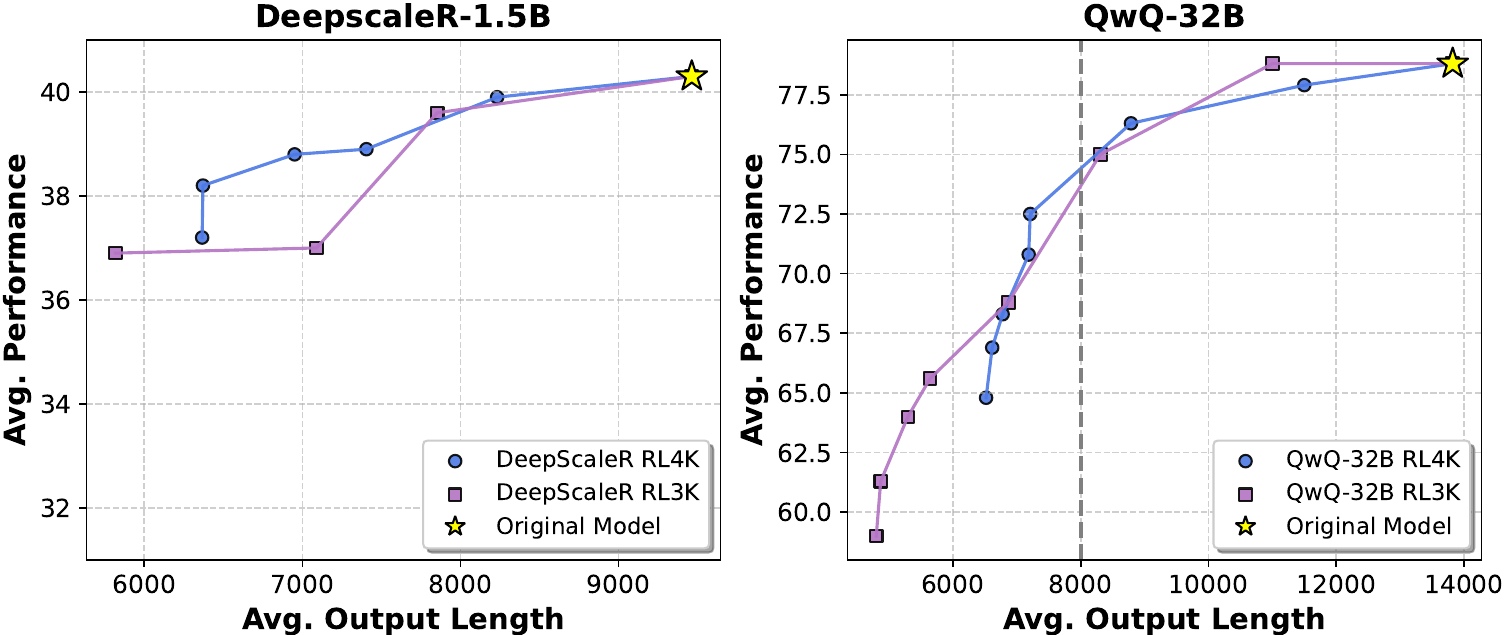}
\caption{
Performance-length trade-off along the length pruning training process evaluated on the AIME24 dataset. The generation length drops quickly in the early stages of training with minor performance drop.
}
\vspace*{-0.2in}
\label{fig: tradeoff}
\end{figure}

\paragraph{Performance-length trade-off along the pruning process.}
We observe that the experiment results in Table~\ref{tab: main_exp} indicates a positive correlation between the reasoning length and model performance on different benchmarks. To better understand the trade-off between reasoning length and model performance, we visualize how these two change during the length pruning training process on the AIME24 dataset (see Figure~\ref{fig: tradeoff}). During training, we evaluate the model every 20 steps and record both the performance and the average output length. We then identify the most effective checkpoints (the ones at the frontier of performance-length trade-off),  defined as the checkpoints that achieve the best performance among all the checkpoints whose average output length is shorter than themselves.
This gives us a clear picture of the best performance the model can reach at each length range, helping us visualize the efficiency boundary of the pruning process.

We highlight the following observations:
\textbf{First}, mild length pruning leads to strong efficiency gains with minimal performance drop. When reasoning length reduction is moderate, there exists a favorable trade-off: significant token reductions can be achieved with only minor drops in performance. For example, we can reduce the average generation length of \texttt{QwQ-32B} from 14K to 8K while maintaining its performance close to the original model.
\textbf{Second}, for the \texttt{QwQ-32B} model, we observe a distinct critical threshold (marked by the vertical dotted line in Figure~\ref{fig: tradeoff}). Beyond this threshold, further reducing the reasoning length causes a sharp and significant drop in performance.
We also find enforcing 4k length limit leads to even worse results than 3k length limit on \texttt{QwQ-32B}. This drop partially explains why iterative pruning performs worse on \texttt{QwQ-32B} than on smaller models like 1.5B.
While optimizing hyperparameters and the training data for \texttt{QwQ-32B} may lead to better performance, we leave the exploration to future work due to the heavy computation cost.

\begin{figure*}[t]
\centering
    \subfigure
    {          
        \centering
        \includegraphics[height=.32\linewidth]{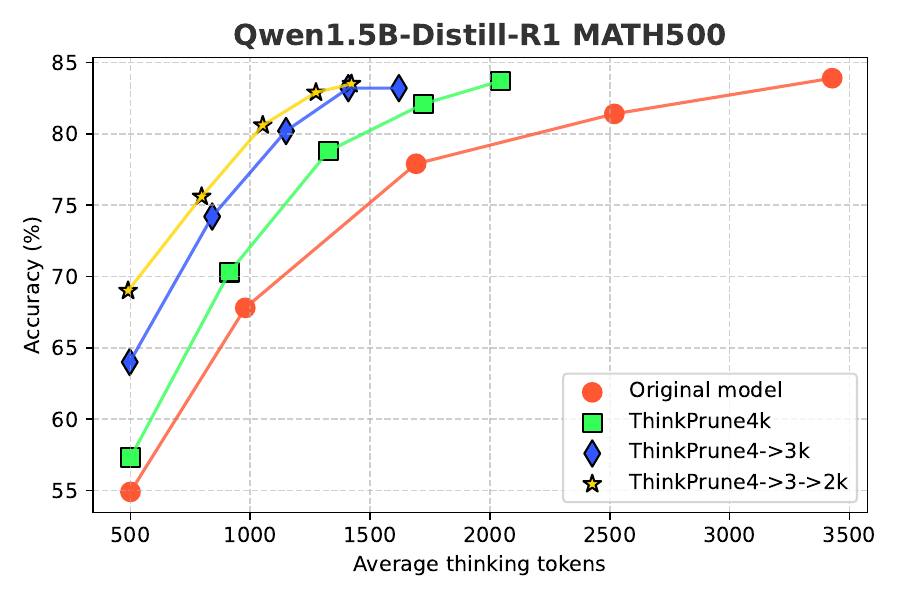}
    }
    \subfigure
    {          
        \centering
        \includegraphics[height=.32\linewidth]{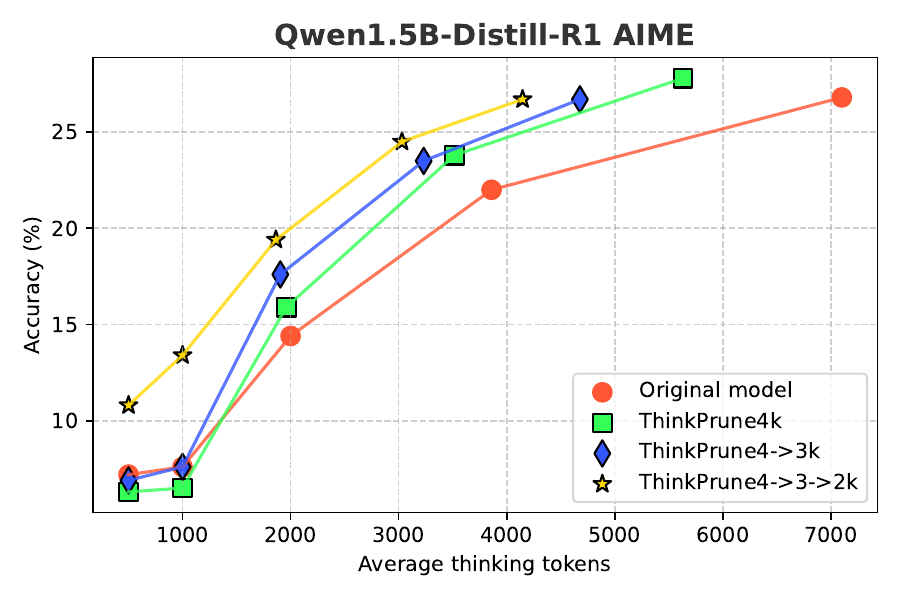}
    }
    \vspace{-5mm}
    \\
    \subfigure
    {          
        \centering
        \includegraphics[height=.32\linewidth]{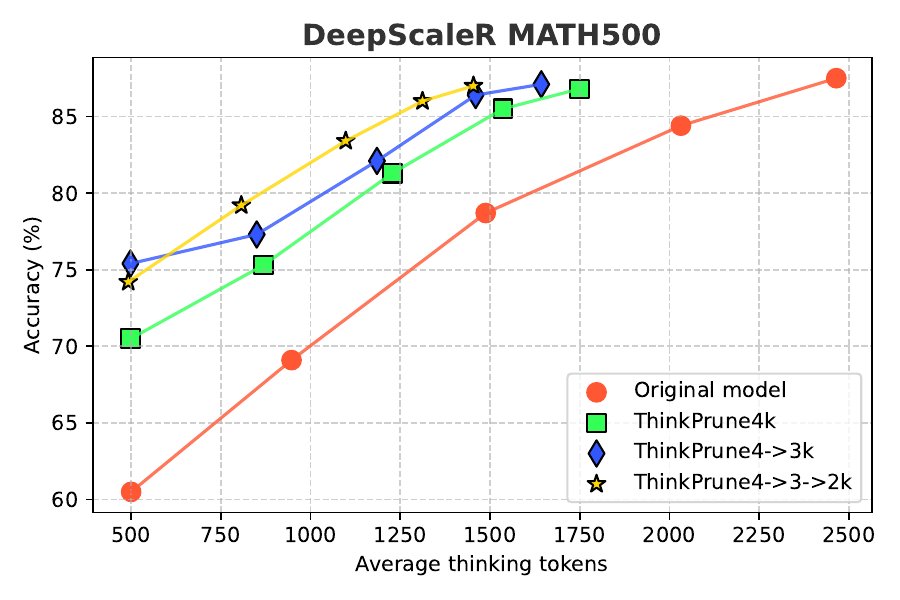}
    }
    \subfigure
    {          
        \centering
        \includegraphics[height=.32\linewidth]{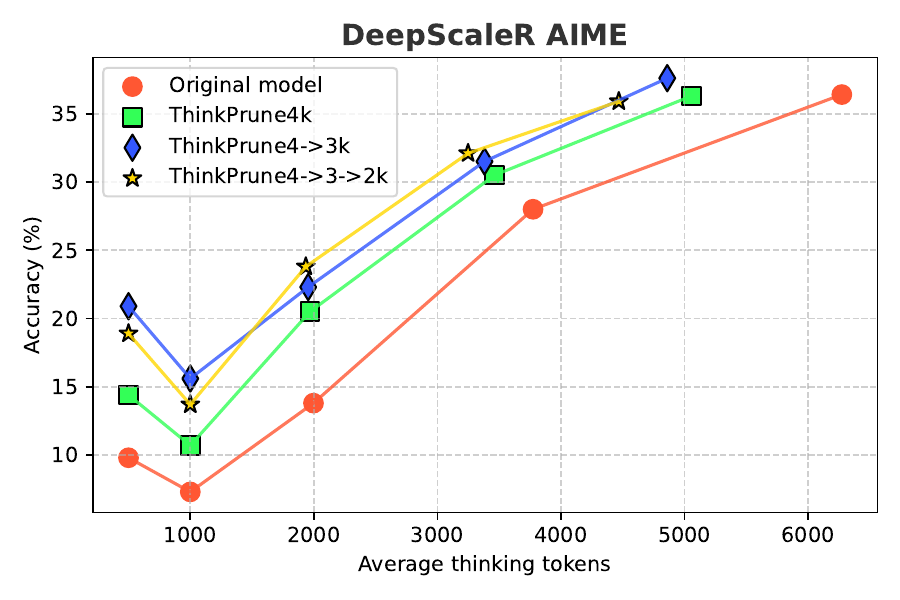}
    }
    \vspace{-5mm}
    \caption{Inference-time thinking length vs. performance trade-off for different long-CoT LLMs with \alg after applying budget-forcing.}
    \label{fig: comp_budget_forcing}
    \vspace*{-0.2in}
\end{figure*}

\paragraph{Inference-time trade-off with budget-forcing.} 
We further study how different long-CoT LLMs perform when applying the budget-forcing method~\citep{muennighoff2025s1}. Detailed budget-forcing prompt can be found in Appendix~\ref{app: implementation}.
Figure~\ref{fig: comp_budget_forcing} shows the trade-off between performance and output length on the Math-500 dataset (left) and the AIME-24 dataset (right), comparing models before and after applying \alg with budget-forcing. We highlight two key observations:
\textbf{First}, \alg significantly improves performance under a limited thinking token budget. For example, \alg-4k$\rightarrow$3k$\rightarrow$2k consistently outperforms the original model when using the same number of thinking tokens. On Math-500, it reaches similar accuracy while using only about 50\% of the original thinking tokens for \textbf{\texttt{Qwen1.5B-Distill-R1}}. This suggests that \alg helps the model think more efficiently and make better use of a limited token budget.
\textbf{Second}, \alg reduces more thinking tokens on easier Math-500 problems than on AIME-24. This suggests that redundancy in reasoning is related to problem difficulty and that length pruning can adaptively remove unnecessary thinking for questions at different difficulties.

\subsection{Reasoning Behavior Analaysis}
In this section, we take a closer look at the reasoning behavior of long-COT LLMs after length pruning. We focus on answering the following two questions:
1. How does the reasoning behavior of long-COT LLMs change, and what gets removed most after pruning?
2. How does the pruning affect the readability of model-generated reasoning?

\begin{wrapfigure}{h}{0.6\textwidth}
\vspace{-2mm}
\resizebox{0.97\linewidth}{!}{
    \centering
    \includegraphics[width=\linewidth, height=!]{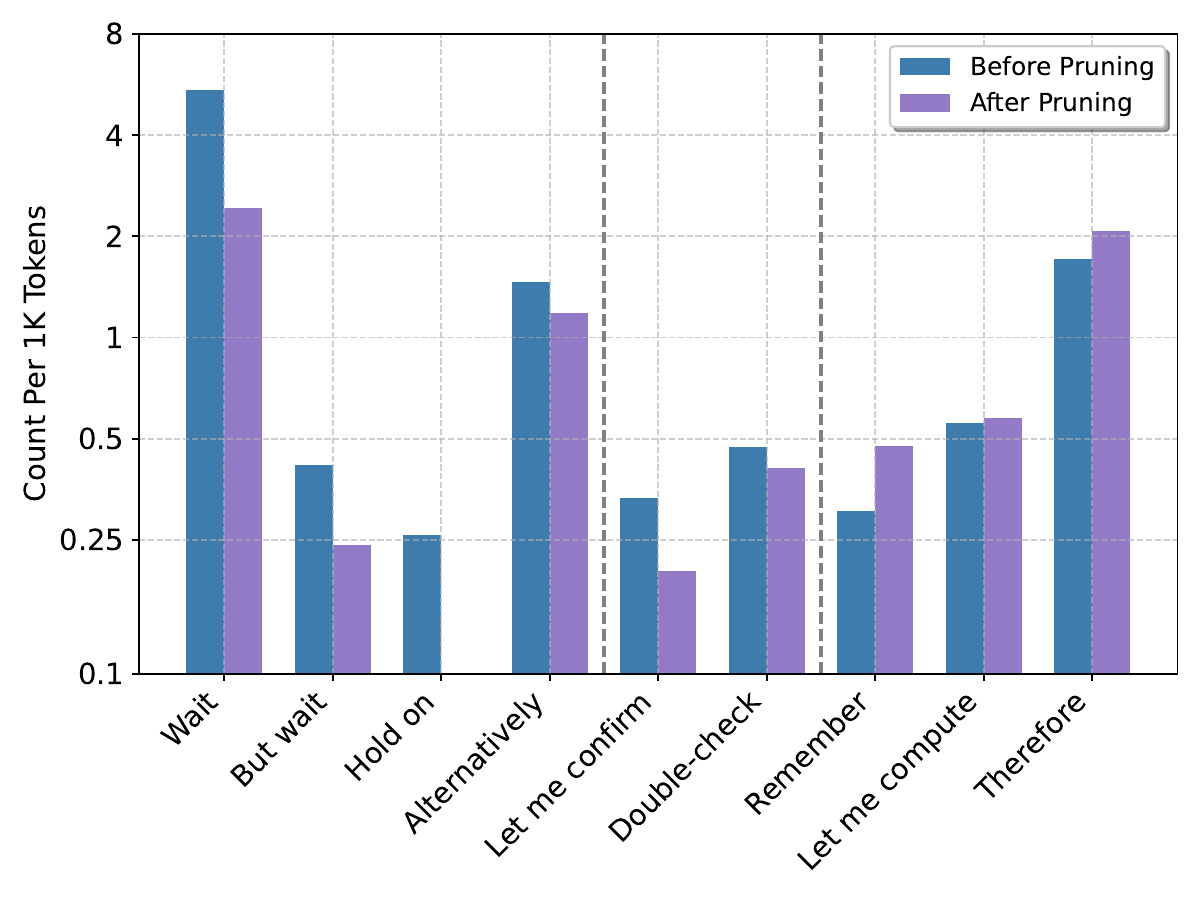}
}
\vspace{-3mm}
\caption{Reasoning-related keyword frequency change before and after length pruning.} 
\label{fig: freq_change}
\end{wrapfigure}

\paragraph{Reasoning-related keyword frequency change.} 
Figure~\ref{fig: freq_change} illustrates the frequency of specific reasoning-related keywords per response in the MATH500 dataset using the \texttt{DeepSeek-R1-Distill-Qwen-1.5B} model, both before and after applying our pruning method. Specifically, we count the number of occurrences of each keywords within the responses of LLMs and then normalize the count by the number of tokens in the responses, which gives us the frequency of each word in 1000 tokens.

The figure is divided into three sections. The left section contains phrases that signal hesitations or self-corrections, which we find undergo a significant drop in frequency after pruning. This indicates that the model would hesitate less. The middle section contains phrases that signal self-verification, which undergo a slight drop in frequency after pruning. This indicates that the model would sometimes skip the self-verification steps, which may sacrifice a little bit of accuracy, to save the token length. Finally, the right section contains phrases that signal core computation and reasoning, which have a slight increase in frequency despite the overall reduction in length. This indicates that the model would maintain the reasoning process, because that is the core process that contributes to the final solution.

\begin{figure}[t]
    \vspace{-3mm}
    \centering
    \includegraphics[width=0.85\linewidth]{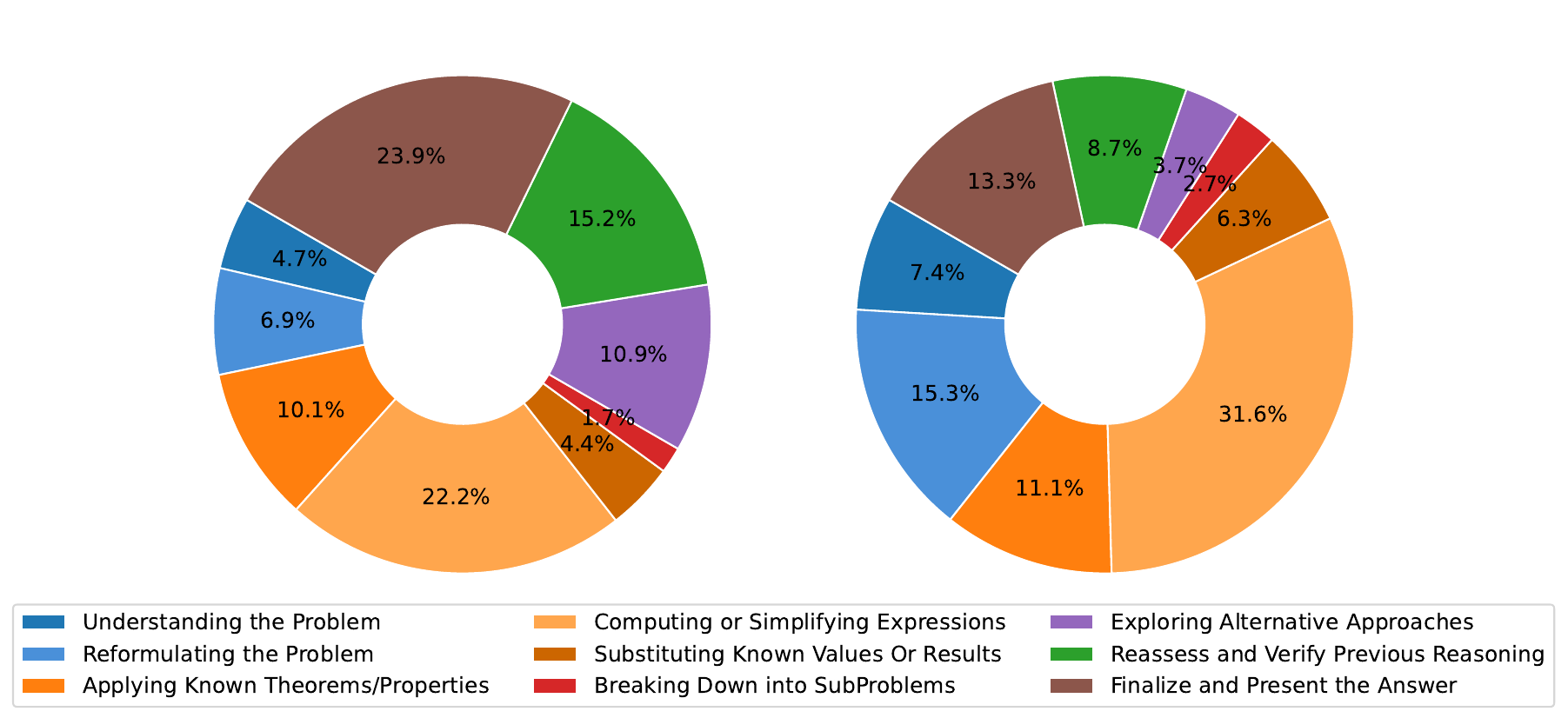}
    \caption{Reasoning behavior change before (left) and after (right) length pruning.}
    \label{fig:problem-solving}
    \vspace*{-0.2in}
\end{figure}

\paragraph{Reasoning behavior change.} 
We further analyze the change in reasoning behaviors that the model utilizes to solve a problem. Specifically, we first prompt GPT-4o to summarize 9 frequent problem-solving phases, such as `understanding the problem' and `applying known theorems/properties' (complete list in Appendix \ref{app: high_level_phase_analysis}). Then, for each model-generated solution, we prompt GPT-4o to split the output into chunks and label each chunk with one of these phases. We measure the number of reasoning steps in each phase by counting segments separated by double newlines (``\textbackslash n \textbackslash n''). Figure~\ref{fig:problem-solving} shows the distribution of these reasoning steps before and after pruning. As can be observed, the model would spend less time on relatively redundant steps, such as `finalize and present the answer', `reassess solution', and `explore alternative approaches'. Meanwhile, we observe an increase in the percentage of core problem-solving steps, including `applying known theorems/properties' and `computing or simplifying expressions'. These observations further confirm that the model would focus more on the core problem-solving steps and save on peripheral steps.

\begin{wrapfigure}{r}{0.35\textwidth}
\centering
\vspace{-3mm}
\resizebox{0.97\linewidth}{!}{
\begin{tabular}{l|c|c}
        \toprule \midrule
        ~ & PPL & Avg. Acc \\ \midrule
        \multicolumn{3}{c}{\textbf{\texttt{DeepSeek-R1-1.5B-Distill}}} \\ \midrule
        Original Model & 1.91 & 82.9 \\ 
        {\alg} & 1.90 & 83.2 \\  \midrule
        \multicolumn{3}{c}{\textbf{\texttt{DeepScaleR-1.5B-Preview}}} \\ \midrule
        Original Model & 1.95 & 88.5 \\ 
        {\alg} & 2.02 & 86.9 \\ \midrule
        \multicolumn{3}{c}{\textbf{\texttt{QwQ-32B}}} \\
        \midrule
        Original Model & 2.37 & 95.1 \\ 
        {\alg} & 2.24 & 93.8 \\ 
        \midrule
        \bottomrule
    \end{tabular}
}
    \vspace{-2mm}
    \captionof{table}{Reasoning trace perplexity on different models.}
    \label{tab:ppl-reasoning}
\end{wrapfigure}

\paragraph{Readability evaluation.} 
One common concern in RLVR is that the heavy RL training may reduce the readability of the model's reasoning, resulting in mixed language or non-readable reasoning trace as shown in R1-Zero model~\citep{deepseek-ai2025deepseekr1}. 
To examine whether \alg would also introduce similar issues, we measure the perplexity of generated reasoning traces on the Math-500 dataset for original model and \alg-4k$\rightarrow$ 3k$\rightarrow$ 2k using Qwen2.5-Math-7B. As shown in Table~\ref{tab:ppl-reasoning}, our pruning method does not significantly affect reasoning readability --- the perplexity remains nearly identical to the original model. 
Figure~\ref {fig:example-trace} in Appendix~\ref{app: example-reasoning} shows an example reasoning trace from the Distill-R1-1.5B LLM before and after pruning. As shown in the example, the original LLM repeatedly checks its previous reasoning multiple times, leading to heavy, redundant reasoning steps. On the contrary, the reasoning trace remains fully readable and focused on solving the problem after pruning, with more efficient problem-solving steps and only one self-verification step.

\section{Conclusion}
In this paper, we proposed {\alg} to reduce the reasoning length of long CoT LLMs. {\alg} introduces a length constraint during RL training, which discards unfinished thoughts and answers when sampling responses. To maintain model performance, we apply an iterative pruning strategy that gradually tightens the length limit over multiple rounds training. Experiments show that {\alg} reduces the reasoning length and achieves a strong performance-length trade-off. Further analysis shows that {\alg} effectively removes redundant steps while preserving key reasoning processes.

\section{Acknowledgments}
The work of Bairu Hou, Jiabao Ji, Yujian Liu, and Shiyu Chang was partially supported by National Science Foundation (NSF) Grant IIS-2338252, NSF Grant IIS-2207052, and NSF Grant IIS-2302730. The work of Jacob Andreas is supported by a Sloan Fellowship and the NSF under grant IIS-2238240. The computing resources used in this work were partially supported by the Accelerate Foundation Models Research program of Microsoft.

\bibliography{colm2025_conference}

\begin{thebibliography}{34}
\providecommand{\natexlab}[1]{#1}
\providecommand{\url}[1]{\texttt{#1}}
\expandafter\ifx\csname urlstyle\endcsname\relax
  \providecommand{\doi}[1]{doi: #1}\else
  \providecommand{\doi}{doi: \begingroup \urlstyle{rm}\Url}\fi

\bibitem[AMC(2025)]{AoPS_AIME}
AMC.
\newblock American invitational mathematics examination.
\newblock \url{https://artofproblemsolving.com/wiki/index.php/American_Invitational_Mathematics_Examination}, 2025.

\bibitem[{AMC}(2025)]{maa_amc}
{AMC}.
\newblock American mathematics competitions (amc).
\newblock \url{https://maa.org/student-programs/amc/}, 2025.

\bibitem[Chen et~al.(2025)Chen, Qin, Liu, Peng, Guan, Wang, Hu, Zhou, Gao, and Che]{chen2025towards}
Qiguang Chen, Libo Qin, Jinhao Liu, Dengyun Peng, Jiannan Guan, Peng Wang, Mengkang Hu, Yuhang Zhou, Te~Gao, and Wangxiang Che.
\newblock Towards reasoning era: A survey of long chain-of-thought for reasoning large language models.
\newblock \emph{arXiv preprint arXiv:2503.09567}, 2025.

\bibitem[Chen et~al.(2024)Chen, Xu, Liang, He, Pang, Yu, Song, Liu, Zhou, Zhang, et~al.]{chen2024not}
Xingyu Chen, Jiahao Xu, Tian Liang, Zhiwei He, Jianhui Pang, Dian Yu, Linfeng Song, Qiuzhi Liu, Mengfei Zhou, Zhuosheng Zhang, et~al.
\newblock Do not think that much for 2+ 3=? on the overthinking of o1-like llms.
\newblock \emph{arXiv preprint arXiv:2412.21187}, 2024.

\bibitem[Cui et~al.(2025)Cui, Yuan, Wang, Wang, Li, He, Fan, Yu, Xu, Chen, et~al.]{cui2025prime}
Ganqu Cui, Lifan Yuan, Zefan Wang, Hanbin Wang, Wendi Li, Bingxiang He, Yuchen Fan, Tianyu Yu, Qixin Xu, Weize Chen, et~al.
\newblock Process reinforcement through implicit rewards.
\newblock \emph{arXiv preprint arXiv:2502.01456}, 2025.

\bibitem[DeepSeek-AI(2025)]{deepseek-ai2025deepseekr1}
DeepSeek-AI.
\newblock Deepseek-r1: Incentivizing reasoning capability in llms via reinforcement learning.
\newblock \emph{arXiv preprint arXiv: 2501.12948}, 2025.

\bibitem[Fu et~al.(2024)Fu, Chen, Zhu, Fu, Dai, Qiao, and Zhang]{fu2024efficiently}
Yichao Fu, Junda Chen, Siqi Zhu, Zheyu Fu, Zhongdongming Dai, Aurick Qiao, and Hao Zhang.
\newblock Efficiently serving llm reasoning programs with certaindex.
\newblock \emph{arXiv preprint arXiv: 2412.20993}, 2024.

\bibitem[Gandhi et~al.(2025)Gandhi, Chakravarthy, Singh, Lile, and Goodman]{gandhi2025cognitive}
Kanishk Gandhi, Ayush Chakravarthy, Anikait Singh, Nathan Lile, and Noah~D Goodman.
\newblock Cognitive behaviors that enable self-improving reasoners, or, four habits of highly effective stars.
\newblock \emph{arXiv preprint arXiv:2503.01307}, 2025.

\bibitem[Gao et~al.(2024)Gao, Tow, Abbasi, Biderman, Black, DiPofi, Foster, Golding, Hsu, Le~Noac'h, Li, McDonell, Muennighoff, Ociepa, Phang, Reynolds, Schoelkopf, Skowron, Sutawika, Tang, Thite, Wang, Wang, and Zou]{eval-harness}
Leo Gao, Jonathan Tow, Baber Abbasi, Stella Biderman, Sid Black, Anthony DiPofi, Charles Foster, Laurence Golding, Jeffrey Hsu, Alain Le~Noac'h, Haonan Li, Kyle McDonell, Niklas Muennighoff, Chris Ociepa, Jason Phang, Laria Reynolds, Hailey Schoelkopf, Aviya Skowron, Lintang Sutawika, Eric Tang, Anish Thite, Ben Wang, Kevin Wang, and Andy Zou.
\newblock A framework for few-shot language model evaluation, 07 2024.
\newblock URL \url{https://zenodo.org/records/12608602}.

\bibitem[Geiping et~al.(2025)Geiping, McLeish, Jain, Kirchenbauer, Singh, Bartoldson, Kailkhura, Bhatele, and Goldstein]{geiping2025scaling}
Jonas Geiping, Sean McLeish, Neel Jain, John Kirchenbauer, Siddharth Singh, Brian~R. Bartoldson, Bhavya Kailkhura, Abhinav Bhatele, and Tom Goldstein.
\newblock Scaling up test-time compute with latent reasoning: A recurrent depth approach.
\newblock \emph{arXiv preprint arXiv: 2502.05171}, 2025.

\bibitem[Han et~al.(2015)Han, Pool, Tran, and Dally]{han2015learningweightsconnectionsefficient}
Song Han, Jeff Pool, John Tran, and William~J. Dally.
\newblock Learning both weights and connections for efficient neural networks, 2015.

\bibitem[Hao et~al.(2024)Hao, Sukhbaatar, Su, Li, Hu, Weston, and Tian]{hao2024training}
Shibo Hao, Sainbayar Sukhbaatar, DiJia Su, Xian Li, Zhiting Hu, Jason Weston, and Yuandong Tian.
\newblock Training large language models to reason in a continuous latent space.
\newblock \emph{arXiv preprint arXiv: 2412.06769}, 2024.

\bibitem[He et~al.(2024)He, Luo, Bai, Hu, Thai, Shen, Hu, Han, Huang, Zhang, Liu, Qi, Liu, and Sun]{he2024olympiadbench0}
Chaoqun He, Renjie Luo, Yuzhuo Bai, Shengding Hu, Zhen~Leng Thai, Junhao Shen, Jinyi Hu, Xu~Han, Yujie Huang, Yuxiang Zhang, Jie Liu, Lei Qi, Zhiyuan Liu, and Maosong Sun.
\newblock Olympiadbench: A challenging benchmark for promoting agi with olympiad-level bilingual multimodal scientific problems.
\newblock \emph{Annual Meeting of the Association for Computational Linguistics}, 2024.
\newblock \doi{10.48550/arXiv.2402.14008}.

\bibitem[Hosseini et~al.(2024)Hosseini, Yuan, Malkin, Courville, Sordoni, and Agarwal]{hosseini2024vstar}
Arian Hosseini, Xingdi Yuan, Nikolay Malkin, Aaron Courville, Alessandro Sordoni, and Rishabh Agarwal.
\newblock V-star: Training verifiers for self-taught reasoners.
\newblock \emph{arXiv preprint arXiv:2402.06457}, 2024.

\bibitem[Ji et~al.(2025)Ji, Xu, Liang, Liu, He, Chen, Liu, Wang, Chen, Wang, et~al.]{ji2025first}
Ke~Ji, Jiahao Xu, Tian Liang, Qiuzhi Liu, Zhiwei He, Xingyu Chen, Xiaoyuan Liu, Zhijie Wang, Junying Chen, Benyou Wang, et~al.
\newblock The first few tokens are all you need: An efficient and effective unsupervised prefix fine-tuning method for reasoning models.
\newblock \emph{arXiv preprint arXiv:2503.02875}, 2025.

\bibitem[KimiTeam et~al.(2025)KimiTeam, Du, Gao, Xing, Jiang, Chen, Li, Xiao, Du, Liao, et~al.]{team2025kimi}
KimiTeam, Angang Du, Bofei Gao, Bowei Xing, Changjiu Jiang, Cheng Chen, Cheng Li, Chenjun Xiao, Chenzhuang Du, Chonghua Liao, et~al.
\newblock Kimi k1. 5: Scaling reinforcement learning with llms.
\newblock \emph{arXiv preprint arXiv:2501.12599}, 2025.

\bibitem[Kumar et~al.(2025)Kumar, Roh, Naseh, Karpinska, Iyyer, Houmansadr, and Bagdasarian]{kumar2025overthink}
Abhinav Kumar, Jaechul Roh, Ali Naseh, Marzena Karpinska, Mohit Iyyer, Amir Houmansadr, and Eugene Bagdasarian.
\newblock Overthink: Slowdown attacks on reasoning llms.
\newblock \emph{arXiv e-prints}, pp.\  arXiv--2502, 2025.

\bibitem[Lightman et~al.(2023)Lightman, Kosaraju, Burda, Edwards, Baker, Lee, Leike, Schulman, Sutskever, and Cobbe]{lightman2023math500}
Hunter Lightman, Vineet Kosaraju, Yura Burda, Harri Edwards, Bowen Baker, Teddy Lee, Jan Leike, John Schulman, Ilya Sutskever, and Karl Cobbe.
\newblock Let's verify step by step.
\newblock \emph{arXiv preprint arXiv:2305.20050}, 2023.

\bibitem[Liu \& Zhang(2025)Liu and Zhang]{code-r1}
Jiawei Liu and Lingming Zhang.
\newblock Code-r1: Reproducing r1 for code with reliable rewards.
\newblock 2025.

\bibitem[Luo et~al.(2025)Luo, Tan, Wong, Shi, Tang, Roongta, Cai, Luo, Zhang, Li, Popa, and Stoica]{deepscaler2025}
Michael Luo, Sijun Tan, Justin Wong, Xiaoxiang Shi, William~Y. Tang, Manan Roongta, Colin Cai, Jeffrey Luo, Tianjun Zhang, Li~Erran Li, Raluca~Ada Popa, and Ion Stoica.
\newblock Deepscaler: Surpassing o1-preview with a 1.5b model by scaling rl.
\newblock \url{https://pretty-radio-b75.notion.site/DeepScaleR-Surpassing-O1-Preview-with-a-1-5B-Model-by-Scaling-RL-19681902c1468005bed8ca303013a4e2}, 2025.
\newblock Notion Blog.

\bibitem[Muennighoff et~al.(2025)Muennighoff, Yang, Shi, Li, Fei-Fei, Hajishirzi, Zettlemoyer, Liang, Cand{\`e}s, and Hashimoto]{muennighoff2025s1}
Niklas Muennighoff, Zitong Yang, Weijia Shi, Xiang~Lisa Li, Li~Fei-Fei, Hannaneh Hajishirzi, Luke Zettlemoyer, Percy Liang, Emmanuel Cand{\`e}s, and Tatsunori Hashimoto.
\newblock s1: Simple test-time scaling.
\newblock \emph{arXiv preprint arXiv:2501.19393}, 2025.

\bibitem[OpenAI(2024)]{openai2024openai}
OpenAI.
\newblock Openai o1 system card.
\newblock \emph{arXiv preprint arXiv: 2412.16720}, 2024.

\bibitem[{OpenAI}(2024)]{openai_operator_2024}
{OpenAI}.
\newblock Introducing operator, 2024.
\newblock URL \url{https://openai.com/index/introducing-operator/}.
\newblock Accessed: 2025-03-23.

\bibitem[OpenAI(2025)]{openai2025competitive}
OpenAI.
\newblock Competitive programming with large reasoning models.
\newblock \emph{arXiv preprint arXiv: 2502.06807}, 2025.

\bibitem[QwenTeam(2025)]{qwq32b}
QwenTeam.
\newblock Qwq-32b: Embracing the power of reinforcement learning, March 2025.
\newblock URL \url{https://qwenlm.github.io/blog/qwq-32b/}.

\bibitem[Shao et~al.(2024)Shao, Wang, Zhu, Xu, Song, Bi, Zhang, Zhang, Li, Wu, and Guo]{shao2024deepseekmath}
Zhihong Shao, Peiyi Wang, Qihao Zhu, Runxin Xu, Junxiao Song, Xiao Bi, Haowei Zhang, Mingchuan Zhang, Y.~K. Li, Y.~Wu, and Daya Guo.
\newblock Deepseekmath: Pushing the limits of mathematical reasoning in open language models.
\newblock \emph{arXiv preprint arXiv: 2402.03300}, 2024.

\bibitem[Shen et~al.(2025)Shen, Yan, Zhang, Hu, Du, and He]{shen2025codi0}
Zhenyi Shen, Hanqi Yan, Linhai Zhang, Zhanghao Hu, Yali Du, and Yulan He.
\newblock Codi: Compressing chain-of-thought into continuous space via self-distillation.
\newblock \emph{arXiv preprint arXiv: 2502.21074}, 2025.

\bibitem[Sheng et~al.(2024)Sheng, Zhang, Ye, Wu, Zhang, Zhang, Peng, Lin, and Wu]{sheng2024verl}
Guangming Sheng, Chi Zhang, Zilingfeng Ye, Xibin Wu, Wang Zhang, Ru~Zhang, Yanghua Peng, Haibin Lin, and Chuan Wu.
\newblock Hybridflow: A flexible and efficient rlhf framework.
\newblock \emph{arXiv preprint arXiv: 2409.19256}, 2024.

\bibitem[Sui et~al.(2025)Sui, Chuang, Wang, Zhang, Zhang, Yuan, Liu, Wen, Chen, Hu, et~al.]{sui2025stop}
Yang Sui, Yu-Neng Chuang, Guanchu Wang, Jiamu Zhang, Tianyi Zhang, Jiayi Yuan, Hongyi Liu, Andrew Wen, Hanjie Chen, Xia Hu, et~al.
\newblock Stop overthinking: A survey on efficient reasoning for large language models.
\newblock \emph{arXiv preprint arXiv:2503.16419}, 2025.

\bibitem[Wang et~al.(2025)Wang, Liu, Xu, Liang, Chen, He, Song, Yu, Li, Zhang, et~al.]{wang2025thoughts}
Yue Wang, Qiuzhi Liu, Jiahao Xu, Tian Liang, Xingyu Chen, Zhiwei He, Linfeng Song, Dian Yu, Juntao Li, Zhuosheng Zhang, et~al.
\newblock Thoughts are all over the place: On the underthinking of o1-like llms.
\newblock \emph{arXiv preprint arXiv:2501.18585}, 2025.

\bibitem[Ye et~al.(2025)Ye, Huang, Xiao, Chern, Xia, and Liu]{ye2025limo}
Yixin Ye, Zhen Huang, Yang Xiao, Ethan Chern, Shijie Xia, and Pengfei Liu.
\newblock Limo: Less is more for reasoning.
\newblock \emph{arXiv preprint arXiv: 2502.03387}, 2025.

\bibitem[Zelikman et~al.(2024)Zelikman, Harik, Shao, Jayasiri, Haber, and Goodman]{zelikman2024quiet}
Eric Zelikman, Georges~Raif Harik, Yijia Shao, Varuna Jayasiri, Nick Haber, and Noah Goodman.
\newblock Quiet-star: Language models can teach themselves to think before speaking.
\newblock In \emph{First Conference on Language Modeling}, 2024.

\bibitem[Zeng et~al.(2025{\natexlab{a}})Zeng, Huang, Liu, Liu, He, Ma, and He]{zeng2025simplerl0zoo0}
Weihao Zeng, Yuzhen Huang, Qian Liu, Wei Liu, Keqing He, Zejun Ma, and Junxian He.
\newblock Simplerl-zoo: Investigating and taming zero reinforcement learning for open base models in the wild.
\newblock \emph{arXiv preprint arXiv: 2503.18892}, 2025{\natexlab{a}}.

\bibitem[Zeng et~al.(2025{\natexlab{b}})Zeng, Cheng, Yin, Zhou, and Qiu]{zeng2025revisiting}
Zhiyuan Zeng, Qinyuan Cheng, Zhangyue Yin, Yunhua Zhou, and Xipeng Qiu.
\newblock Revisiting the test-time scaling of o1-like models: Do they truly possess test-time scaling capabilities?
\newblock \emph{arXiv preprint arXiv: 2502.12215}, 2025{\natexlab{b}}.

\end{thebibliography}
\bibliographystyle{colm2025_conference}

\appendix
\section{Appendix}

\subsection{Implementation Details}
\label{app: implementation}

\paragraph{System prompt used for training.} The system prompt for DeepSeek-R1-Distill-Qwen-1.5B and DeepScaleR during training is shown in Table~\ref{tab:system_prompt} to align with the original RL training of DeepSeek-R1. For QwQ-32B, we use a much similar prompt, ``you are a helpful assistant. Your output should be within \{N\} tokens.''.
\begin{table}[h]
    \centering
    \resizebox{0.97\linewidth}{!}{
    \begin{tabular}{l}
    \toprule
    A conversation between User and Assistant. The user asks a question, and the Assistant solves it. \\ The assistant first thinks about the reasoning process in the mind and then provides the user with\\the answer.
    The reasoning process and answer are enclosed within $<$think$>$ $<$/think$>$ and \\ $<$answer$>$  $<$/answer$>$ tags, respectively, i.e.,  $<$think$>$ reasoning process here  $<$/think$>$ \\  $<$answer$>$ answer here  $<$/answer$>$. \textcolor{red}{The output of the assistant should be within \{N\} tokens.}\\
     \bottomrule
    \end{tabular}
    }
    \caption{Template for DeepSeek-R1-Distill-Qwen-1.5B and DeepScaleR. \{N\} will be replaced with the length limit for training (\emph{e.g.}, 2000 and 4000). }
    \label{tab:system_prompt}
\end{table}

\paragraph{Implementation of budget forcing.} We follow the official implementation of budget forcing in S1~\cite{muennighoff2025s1} and made minor changes. Since the original implementation is coupling with the \texttt{lm-harness-eval} framework~\citep{eval-harness}, we revise the code to remove such dependency. Also, to stop the thinking process of the LLM, we append ``$<$/think$>$\textbackslash n\textbackslash n**Final Answer:**\textbackslash n\textbackslash n'' to the generation of the LLM instead of ``$<|$im\_start$|>$answer\textbackslash nFinal Answer:'' used by the S1 model. This is because we empirically find the DeepSeek-R1-Distill-Qwen-1.5B model typically summarize its final answer starting with ``**Final Answer:**\textbackslash n\textbackslash n''. Finally, we sample multiple responses to reduce the variance in model performance instead of using greedy decoding as the original implementation.

\subsection{Analyze the Reasoning Behavior Change}
\label{app: high_level_phase_analysis}
We use GPT-4o to analyze the reasoning behavior of LLMs by segmenting their long-form solutions into high-level problem-solving phases. The prompt used for this task is shown in Figure~\ref{tab: prompt_high_level}. Since the model-generated solutions are very long, we only require GPT-4o to output of the first step and last step in each chunk to represent that chunk. Since the model-generated solutions are often very long, we ask GPT-4o to return only the first and last reasoning steps of each phase, which serve as markers to define the boundaries of each chunk.

To align these chunks with the original model output, we use string matching to locate the start and end positions of each chunk in the raw text. Within each matched chunk, we estimate the number of reasoning steps by counting the number of ``\textbackslash n \textbackslash n'' delimiters. This gives us a step-level breakdown of how much reasoning is devoted to each phase.

\begin{table}[h]
    \centering
    \resizebox{0.97\linewidth}{!}{
    \begin{tabular}{l}
    \toprule
    \#\#\# **Task Description**\\
    \\
    Given a mathematical question and its detailed solution, the task is to segment the solution into high-level\\problem-solving phases. The goal is to group consecutive steps into meaningful phases and output only \\the start and end steps of each phase.\\
    \\    
    Note: Each **reasoning step** in the solution is separated by a **double line break ("\textbackslash n\textbackslash n")**.
    \\
    $------$\\
    \#\#\# **Requirements** \\
    \\
    1. **Segment the full solution into distinct problem-solving phases** based on logical progression.\\
    2. **Each phase should have a start and an end step**.\\
    3. **A phase can appear multiple times** in different parts of the solution.\\
    4. **The order of phases is flexible**—they can appear in any logical sequence depending on the nature of the solution.\\
    5. **Only the first and last steps of each phase should be output**, reducing redundancy.\\
    \\    
    $------$\\

\#\#\# **High-Level Problem-Solving Phases**\\
\\
Each step in the solution should belong to one of the following **ten high-level phases**:\\
\\
1. **Understanding the Problem**: Identifying given data, definitions, and the goal.\\
2. **Reformulating the Problem**: Changing variables, rewriting expressions, or restructuring sums.\\
3. **Applying Known Theorems/Properties**: Using standard formulas, identities, or mathematical principles.\\
4. **Breaking Down into Subproblems**: Decomposing the problem into manageable components.\\
5. **Computing or Simplifying Expressions**: Performing algebraic manipulation or numerical evaluation.\\
6. **Substituting Known Values or Results**: Using precomputed values or standard mathematical constants.\\
7. **Reassess and Verify Local Steps**: Checking for errors or inconsistencies within a small part of the solution.\\
8. **Reassess the Whole Solution**: Reviewing the entire solution for logical correctness and consistency.\\
9. **Exploring Alternative Approaches**: Considering different methods to solve the problem.\\
10. **Finalize and Present the Answer**: Writing the final result and ensuring clarity.\\
$------$ \\
\#\#\# **Output Format**\\
The output should consist of **multiple phases**, each represented in the following format:\\
\\
\texttt{```}\\
{[} Phase X {]}: \{Phase Name\}\\
{[}Start{]}: \{Text of first step in the phase\}\\
{[}End{]}: \{Text of last step in the phase\}\\
\texttt{```}\\
\\
Where:\\
\\
- **Phase X** represents the index of the phase (e.g., Phase 1, Phase 2, etc.).\\
- **Phase Name** is one of the ten high-level categories.\\
- **Start** is the first step of the phase.\\
- **End** is the last step of the phase.\\
\\
\#\#\#\# **Example Output**\\
\\
```\\
{[}Phase 1{]}: Understanding the Problem\\
{[}Start{]}: {[}Text of step 1{]}\\
{[}End{]}: {[}Text of step 3{]}\\
\\
{[}Phase 2{]}: Reformulating the Problem\\
{[}Start{]}: {[}Text of step 4{]}\\
{[}End{]}: {[}Text of step 5{]}\\
\\
{[}Phase 3{]}: Computing or Simplifying Expressions\\
{[}Start{]}: {[}Text of step X{]}\\
{[}End{]}: {[}Text of step Y{]}\\
...
\\
{[}Phase 4{]}: Finalize and Present the Answer\\
{[}Start{]}: {[}Text of step X{]}\\
{[}End{]}: {[}Text of step Y{]}\\
\texttt{```} \\
    \bottomrule
    \end{tabular}
    }
    \caption{Template for DeepSeek-R1-Distill-Qwen-1.5B and DeepScaleR. \{N\} will be replaced with the length limit for training (\emph{e.g.}, 2000 and 4000). }
    \label{tab: prompt_high_level}
\end{table}

\subsection{Additional Examples}
\label{app: example-reasoning}

\begin{figure}[t]
    \centering
    \includegraphics[width=0.85\linewidth]{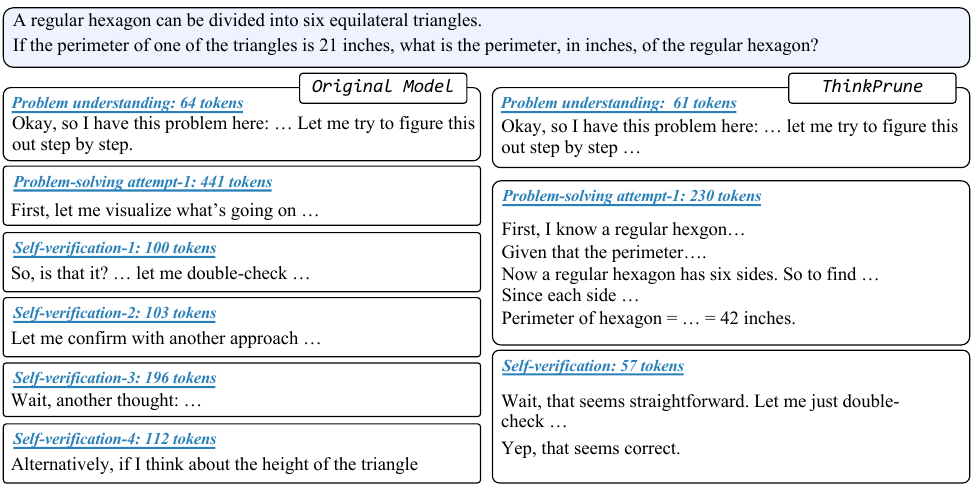}
    \caption{Example reasoning trace of Distill-R1-1.5B before and after length pruning on a Math-500 question.}
    \label{fig:example-trace}
\end{figure}

\begin{figure}[t]
    \centering
    \includegraphics[width=0.85\linewidth]{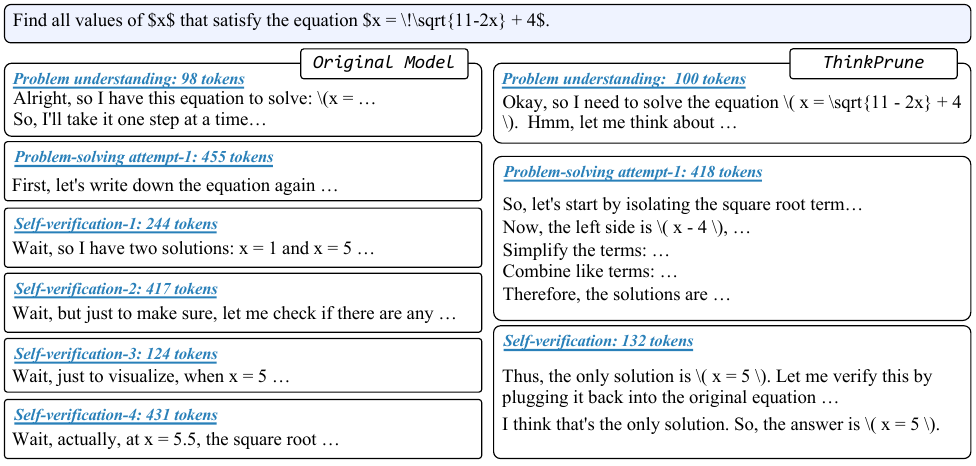}
    \caption{Example reasoning trace of Distill-R1-1.5B before and after length pruning on a Math-500 question.}
    \label{fig:example-trace-2}
\end{figure}

In this section, we include additional reasoning trace examples of Distill-R1-1.5B LLM before and after applying \alg on Math-500 questions. As shown in Figure~\ref{fig:example-trace} and Figure~\ref{fig:example-trace-2}, the original model repeatedly checks its previous reasoning for these simple math questions, leading to many redundant self-reflection steps. On the contrary, our method successfully removes these repeated steps, and helps the LLM focus on the problem solving, while keeping perfect readability.

\end{document}